\documentclass[12pt]{article}
\usepackage[utf8]{inputenc}
\usepackage[T1]{fontenc}
\usepackage{url}
\usepackage{hyperref}

\usepackage[numbers,sort&compress]{natbib}
\usepackage{booktabs}
\usepackage{array}
\usepackage{graphicx}
\usepackage{tabularx}
\usepackage{placeins}
\usepackage{caption}
\usepackage{xcolor}

\begin{document}


\title{MURAD: A Large-Scale Multi-Domain Unified Reverse Arabic Dictionary Dataset}


\author{
    \mdseries Serry Sibaee\textsuperscript{1},
    Yasser Alhabashi\textsuperscript{1},
    Nadia Sibai\textsuperscript{1},
    Yara Farouk\textsuperscript{1},\\
    Adel Ammar\textsuperscript{1}\textsuperscript{*},
    Sawsan AlHalawani\textsuperscript{1},
    Wadii Boulila\textsuperscript{1} \\
    \\
    \textsuperscript{1}Robotics and Internet-of-Things Laboratory (RIOTU),\\ Prince Sultan University, Riyadh 11586, Saudi Arabia \\
    \textsuperscript{*}Corresponding author: \texttt{aammar@psu.edu.sa}
}

\maketitle

\begin{abstract}
Arabic is a linguistically and culturally rich language with a vast vocabulary that spans scientific, religious, and literary domains. Yet, large-scale lexical datasets linking Arabic words to precise definitions remain limited. We present MURAD (Multi-domain Unified Reverse Arabic Dictionary), an open lexical dataset with 96,243 word–definition pairs. The data come from trusted reference works and educational sources. Extraction used a hybrid pipeline integrating direct text parsing, optical character recognition, and automated reconstruction. This ensures accuracy and clarity. Each record aligns a target word with its standardized Arabic definition and metadata that identifies the source domain. The dataset covers terms from linguistics, Islamic studies, mathematics, physics, psychology, and engineering. It supports computational linguistics and lexicographic research. Applications include reverse dictionary modeling, semantic retrieval, and educational tools. By releasing this resource, we aim to advance Arabic natural language processing and promote reproducible research on Arabic lexical semantics.
\end{abstract}

\section{Background \& Summary}

\subsection{\textbf{Motivation and Context}}

One widely recognized use of reverse dictionaries (RD) is mitigating the “tip of the tongue” (TOT) phenomenon~\cite{Brown1966}. This describes the frustrating state where one recalls a concept or meaning but cannot retrieve the exact word. RD systems let users search by meaning. As a result, they help produce more accurate language and support precise term selection in writing, academic research, and technical communication~\cite{qi-etal-2020-wantwords}. In specialized domains such as law and engineering, RDs map natural-language descriptions to semantically appropriate terminology, enhancing clarity and precision~\cite{Yan2020}.

In Arabic-speaking legal and engineering contexts, this need for precise concept-to-term mapping is especially acute. Legal Arabic combines classical juristic terminology with modern codified language, and the same concept may be expressed using different terms across countries or schools of jurisprudence, increasing the risk of ambiguity or misinterpretation in contracts and regulations~\cite{el2015arabic, Ammar_Legal_2024}. Technical and engineering domains face a related challenge: many modern concepts are borrowed from English or French, leading to competing Arabic translations and inconsistent usage across institutions and textbooks. In both settings, RD systems can help practitioners move from an informal or approximate description of a concept to the most appropriate standardized term, supporting terminological consistency and reducing ambiguity in high-stakes communication.

While RD systems are well-established for languages such as English, French, and Chinese~\cite{Yan2020,qi-etal-2020-wantwords,almeman-etal-2023-3d}, the development of comparable resources for Arabic has lagged significantly~\cite{AlMatham2023}. This is largely due to the language’s morphological richness, diglossia, and orthographic ambiguity, all of which complicate consistent lexical annotation and dataset creation. As a result, Arabic natural language processing (NLP) has lacked large-scale, semantically aligned datasets that connect words to their formal definitions, which are essential for meaning-based applications such as semantic retrieval and word-sense disambiguation~\cite{Alshammari2024,Sibaee2024}.

To address this gap, we present MURAD (\textbf{M}ulti-domain \textbf{U}nified \textbf{R}everse \textbf{A}rabic \textbf{D}ictionary), named after the Arabic term for ``intent'' or ``that which is sought.'' This dataset consists of 96,243 Arabic \textit{(definition, word, source)} triplets, carefully curated from contemporary Arabic dictionaries. Each definition was refined using eight formal lexicographic standards to ensure clarity, consistency, and semantic precision~\cite{Sibaee2025}. The dataset was designed to support both human interpretability and computational modeling, providing a reliable, linguistically grounded resource for Arabic semantic technologies.

To clearly outline the scope and significance of our contribution, we summarize the main aspects of this work as follows:

\begin{itemize}
\item \textbf{The largest curated Arabic word-definition dataset}, developed in accordance with established lexicographic standards and validated through expert review.
\item \textbf{Comprehensive multi-domain coverage}, encompassing classical Arabic, linguistics, Islamic studies, as well as scientific and technical terminology.
\item \textbf{An open and fully reproducible release}, providing complete data processing pipelines and a publicly available dataset.
\end{itemize}

\subsection{\textbf{Dataset Description and Usage}}

Each record in the dataset includes a target Arabic word and its formal definition. This format enables direct use in modeling and evaluation pipelines for tasks such as semantic retrieval, definition generation, and embedding validation.

Beyond reverse dictionary applications, the dataset has broad potential for Arabic NLP research. It can support semantic retrieval systems addressing “tip-of-the-tongue” scenarios \cite{Brown1966,AlMatham2023}, enable word-sense disambiguation through lexically distinct definitions \cite{Alshammari2024}, and provide training data for definition modeling, where models learn to generate or interpret dictionary-like glosses \cite{Mickus2022}. The structure and linguistic consistency of the dataset also make it suitable for embedding evaluation, interpretable model analysis, and cross-lingual semantic alignment studies \cite{jincy2014survey}.

The creation of this dataset builds upon and extends previous efforts in Arabic lexical semantics. The KSAA-RD Shared Task introduced by Al-Matham \textit{et al.}~\cite{AlMatham2023} provided one of the earliest Arabic resources for reverse-dictionary modeling, offering a medium-scale collection of word--definition pairs with embedding representations. However, KSAA-RD contains only 58,000 words, does not apply formal lexicographic guidelines \cite{Sibaee2024} 
, and is limited in domain breadth. The related KSAA-CAD dataset focuses on concept--attribute descriptions rather than dictionary-style definitions and is correspondingly smaller and narrower in scope. In contrast, our MURAD dataset comprises 96,243 curated word--definition pairs drawn from 17 major reference works, spans classical, linguistic, Islamic, scientific, and technical domains, and uses structured lexicographic standards to ensure definition precision and semantic consistency. The dataset is fully open and publicly released, supporting reproducible research in Arabic semantic technologies.

In the multilingual context, the SemEval-2022 Task 1: CODWOE \cite{Mickus2022} explored the bidirectional mapping between dictionary definitions and word embeddings across English, French, Spanish, and Chinese. Our dataset complements this effort by extending the definition–embedding alignment to Arabic, a low-resource, morphologically rich language not included in CODWOE.

Additionally, the Azhary Arabic Lexical Ontology \cite{Ishkewy2014} provides structured semantic relations such as synonymy and hypernymy, supporting ontology-based reasoning. However, Azhary lacks the definitional text required for embedding-based modeling. Our dataset complements it by offering natural-language definitions aligned with target words, enabling data-driven tasks such as semantic retrieval, definition generation, and embedding evaluation. Together, these resources form a stronger foundation for comprehensive Arabic lexical and semantic research. Table \ref{tab:comparison} summarizes the differences between MURAD and some existing  lexical resources.

\begin{table}[h!]
\centering
\renewcommand{\arraystretch}{1.3}
\footnotesize
\caption{Comparison of MURAD with some existing  lexical resources. MURAD demonstrates superior coverage and structured lexicography.}
\begin{tabular}{>{\raggedright\arraybackslash}p{2.1cm} 
                >{\raggedright\arraybackslash}p{2.5cm} 
                >{\raggedright\arraybackslash}p{2.2cm} 
                >{\raggedright\arraybackslash}p{2.5cm} 
                >{\raggedright\arraybackslash}p{2.3cm}}
\toprule
\textbf{Attribute} & \textbf{MURAD [Ours]} & \textbf{KSAA-CAD \cite{AlMatham2023}} & \textbf{SemEval-2022 CODWOE \cite{Mickus2022}} & \textbf{Azhary \cite{Ishkewy2014}} \\
\midrule
Number of Entries & \textbf{96,243 curated } & 58,010 Arabic; 4,355 English & Different entries per language  & 26,195 words / 13,328 synsets \\\midrule
Domains & \textbf{Classical and domain specific} & General & General & General Arabic \\\midrule
Sources & \textbf{17 curated sources} & Single dictionary & Multiple dictionaries \& embeddings & Quran \& manual semantics \\\midrule
Lexicographic standard & \textbf{Structured} & LMF inherited & None & None \\\midrule
Access & \textbf{Fully open} & Task participants & Publicly described & Conditional \\\midrule
Arabic-English Embedding & \textbf{None} & Partial & None & Primarily Arabic \\\midrule
Multilingual & \textbf{Arabic (Classical \& MSA)} & Partial Arabic-English & English, French, Spanish, Russian, Italian   & Primarily Arabic \\
\bottomrule
\end{tabular}
\label{tab:comparison}
\end{table}

\section{Methods}

\subsection{Input Data}

The Arabic Reverse Dictionary (MURAD) dataset was compiled from publicly available Arabic reference works and specialized glossaries across multiple academic domains, including Islamic studies, linguistics, mathematics, chemistry, physics, psychology, and engineering. Primary sources included classical works such as \textit{Al-Kafawi’s Dictionary of Universals}~\cite{kafawi_kulliyat_1998}, \textit{Al-Jurjani’s Book of Definitions}~\cite{jurjani_taarifat_1983}, and several domain-specific dictionaries of scientific terminology which are listed in table \ref{tab:sources}. All input materials were obtained from open-access Arabic digital libraries and repositories that permit non-commercial research use, ensuring broad coverage and accessibility.

Table~\ref{tab:sources} lists the Arabic dictionaries and terminological works used to construct the dataset, including the English translation of each title and the number of extracted definitions. Together, these sources cover a broad lexical and disciplinary range, ensuring balanced representation of classical and modern Arabic terminology.

\begin{table}[h!]
\centering
\renewcommand{\arraystretch}{1.3}
\footnotesize
\begin{tabular}{c p{12.1cm} r}
\toprule
\textbf{Ref ID} & \textbf{English Translation of the Title} & \textbf{Definitions} \\
\midrule

1 & Al-Kafawi's Dictionary of Universals \cite{kafawi_universals} & 14,476 \\

2 & Al-Jurjani's Book of Definitions \cite{jurjani_definitions} & 1,399 \\

3 & Dictionary of Chemistry Terms \cite{chemistry_terms} & 4,468 \\

4 & Dictionary of Machine Learning Terms \cite{ml_dl_terms} & 1,758 \\

5 & Dictionary of Mathematical Terms \cite{math_terms} & 7,808 \\

6 & Dictionary of Physics Terms \cite{physics_terms} & 5,081 \\

7 & Dictionary of Arabic Measurement Terms \cite{measurement_terms} & 550 \\

8 & Dictionary of Psychology Terms \cite{psychology_terms} & 4,157 \\

9 & Dictionary of Mechanical Engineering Terms \cite{mechanical_engineering_terms} & 1,569 \\

10 & Book of Terminology in Arabic Sciences \cite{arabic_sciences} & 13,181 \\

11 & Encyclopedic Dictionary of Applied Linguistics Terms \cite{linguistic_definitions} & 13,350 \\

12 & Dictionary of Islamic Jurisprudence Terms \cite{fiqh_terms} & 9,964 \\

13 & Dictionary of Electrical, Electronic, and Communication Engineering Terms \cite{electrical_engineering_terms} & 1,422 \\

14 & Dictionary of Scholars' Terminology \cite{dastur_al_ulama} & 7,907 \\

15 & Encyclopedia of Faith Terminology \cite{faith_encyclopedia} & 4,160 \\

16 & General Terminology Dictionary \cite{terminologyenc_categories} & 3,750 \\

17 & SDAIA Data and Artificial Intelligence Glossary \cite{sdaia_glossary} & 1,243 \\

\bottomrule
\end{tabular}
\caption{Primary Arabic dictionaries and terminological sources used to compile the MURAD dataset, with reference IDs (as included in the dataset), and the total number of definition pairs extracted from each source. Note: The English translations of the sources' titles are included for readability.}
\label{tab:sources}
\end{table}

\subsection{Data Preprocessing}

Most of the textual data was available in digital PDF or structured HTML formats. However, some sources were only accessible as printed volumes and needed to be converted into digital formats. Therefore, high-resolution scans were pre-processed using Mistral OCR \cite{mistral2025ocr} to produce machine-readable text~\cite{nacar2025sard}. Then, semantic text extraction and structuring were  carried out using GPT-4o \cite{islam2025gpt} to ensure the clarity of definitions and the consistency of textual structure (to extract structured word and definition pairs from the given text).

Arabic text was extracted from PDF and HTML sources using automated scripts designed to preserve character order and diacritics. For digitalized PDF documents, page-level extraction routines parsed each file line by line, and regular expression filters were applied to retain valid Arabic Unicode ranges while removing non-textual elements such as pagination, tables, and headers. Similar semi-structured extraction approaches have been used in previous Arabic NLP corpora \cite{AlMatham2023,Alshammari2024}.

Following text extraction, lines were segmented and joined into continuous definition entries to ensure accurate pairing between each lexical term and its explanatory definition. The preprocessing steps included normalization of Arabic character variants and unification of punctuation. Duplicate and incomplete entries were automatically detected and excluded to maintain internal consistency. The resulting term–definition pairs were stored in a structured tabular format suitable for integration into machine learning pipelines. This preprocessing approach follows best practices in Arabic text normalization and definition-based modeling \cite{Sibaee2024, Antoun2021}.

\begin{figure}[h]
  \centering
  \includegraphics[width=\textwidth]{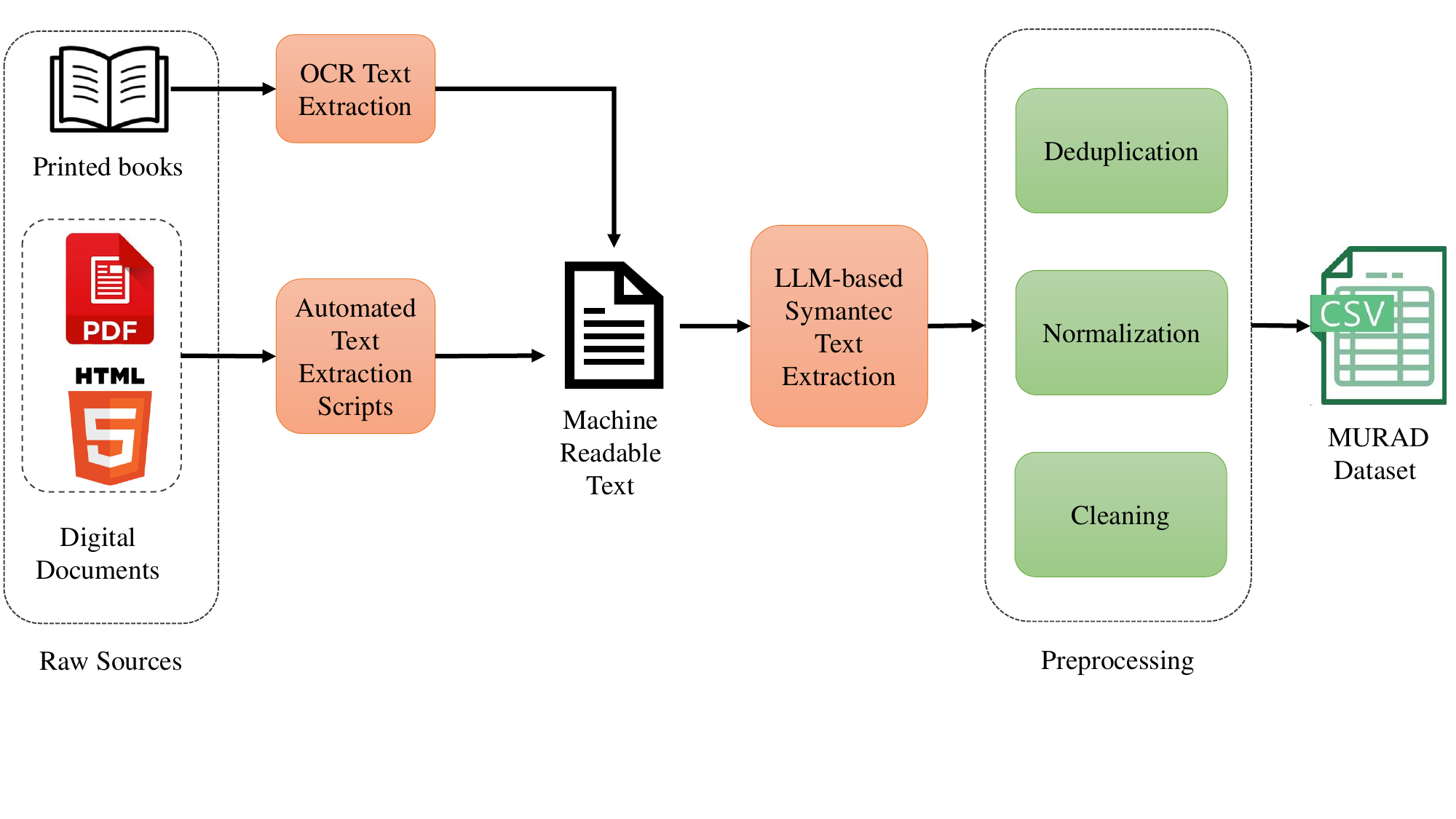}
  \caption{Overview of the Arabic Reverse Dictionary data processing workflow.}
  \label{fig:workflow}
\end{figure}

A high-level overview of the complete data processing workflow is illustrated in Figure~\ref{fig:workflow}. It summarizes each stage of the pipeline, beginning with raw lexical sources and progressing through extraction, normalization, and correction, culminating in the creation of the final dataset, definition embeddings, and validation statistics.

\subsection{Dataset Composition }

After extraction and cleaning, the verified corpus consisted of 96,243 curated term–definition pairs. Each record includes both the textual word and its definition, enabling downstream modeling and semantic evaluation.

All processed data were merged, deduplicated, and exported in UTF-8 encoded CSV format with three standardized fields: \texttt{Word}, \texttt{Definition}, and \texttt{Reference}. Each field corresponds to a column in the dataset which are explained in table \ref{tab:attributes}. This structured release supports reproducibility and reuse in computational linguistic tasks such as reverse dictionary modeling, semantic retrieval, and definition-based embedding evaluation \cite{Yan2020,Qi2020}.


\section{Data Records}
MURAD dataset provides a comprehensive collection of 96,243 Arabic (Word, Definition, Reference) triplets gathered from trusted open-source Arabic reference works. 
Each entry represents a distinct term and its formal definition written in Classical or Modern Standard Arabic. 
The data are distributed in comma-separated values (CSV) format for ease of access and interoperability with common data analysis and NLP frameworks. 
The dataset is stored as a single plain-text CSV file, comma-delimited, and encoded in UTF-8.

The repository contains the following items:
\begin{itemize}
    \item \textbf{rd\_dataset.csv}: the principal file containing the definitional corpus. 
    \item \textbf{README.md}: A description file of the dataset and the used sources. 
\end{itemize}

\begin{table}[h!]
\centering
\begin{tabular}{p{3cm}p{11cm}}
\toprule
\textbf{Column name} & \textbf{Description} \\
\midrule
\texttt{word} & The target Arabic word or term. \\[4pt]
\texttt{definition} & The formal Arabic definition associated with the word. \\[4pt]
\texttt{ref} & An identifier that refers to the source from which the definition was extracted, as listed in Table~\ref{tab:sources}. \\[4pt]
\bottomrule
\end{tabular}
\caption{Attribute descriptions for the Arabic Reverse Dictionary dataset.}
\label{tab:attributes}
\end{table}

Each record in the dataset corresponds to a single lexical entry and includes the columns shown in Table~\ref{tab:attributes}. In addition to the per-record attributes, Table~\ref{tab:summary} presents an overall summary of the dataset composition and scale. It reports the total number of word--definition pairs, contributing sources, and the distribution across 13 domains: Fiqh (Islamic jurisprudence), Aqidah (Islamic creed), Chemistry, Machine Learning, Mathematics, Measurement, Psychology, Mechanics, Linguistics, Electricity, Terminology, Artificial Intelligence, and General Terms. These domains are mapped to four high-level categories: Islamic, Language, Scientific, and General.

\begin{table}[h!]
\centering
\renewcommand{\arraystretch}{1.2}
\begin{tabular}{l|r}
\toprule
\textbf{Metric} & \textbf{Value} \\
\midrule
Total word--definition pairs & 96,243 \\
Total word tokens (all text) & 1,482,322 \\
Unique vocabulary (corpus-wide) & 95,746 \\
Average definition length & 13.8 words \\
Number of sources & 17\\
Number of domains & 13  \\
Number of categories & 4  \\
\bottomrule
\end{tabular}
\caption{Summary statistics of the MURAD dataset. Note: ``Total word tokens'' counts all word occurrences across the entire dataset (words+definitions); ``Unique vocabulary'' represents distinct word types appearing anywhere in the corpus.}
\label{tab:summary}
\end{table}

To provide a concrete sense of the dataset’s content and range, Table~\ref{tab:samples} presents three representative samples illustrating variation in domain, style, and definition length.

\begin{table}[h!]
\centering
\includegraphics[width=1.1\textwidth]{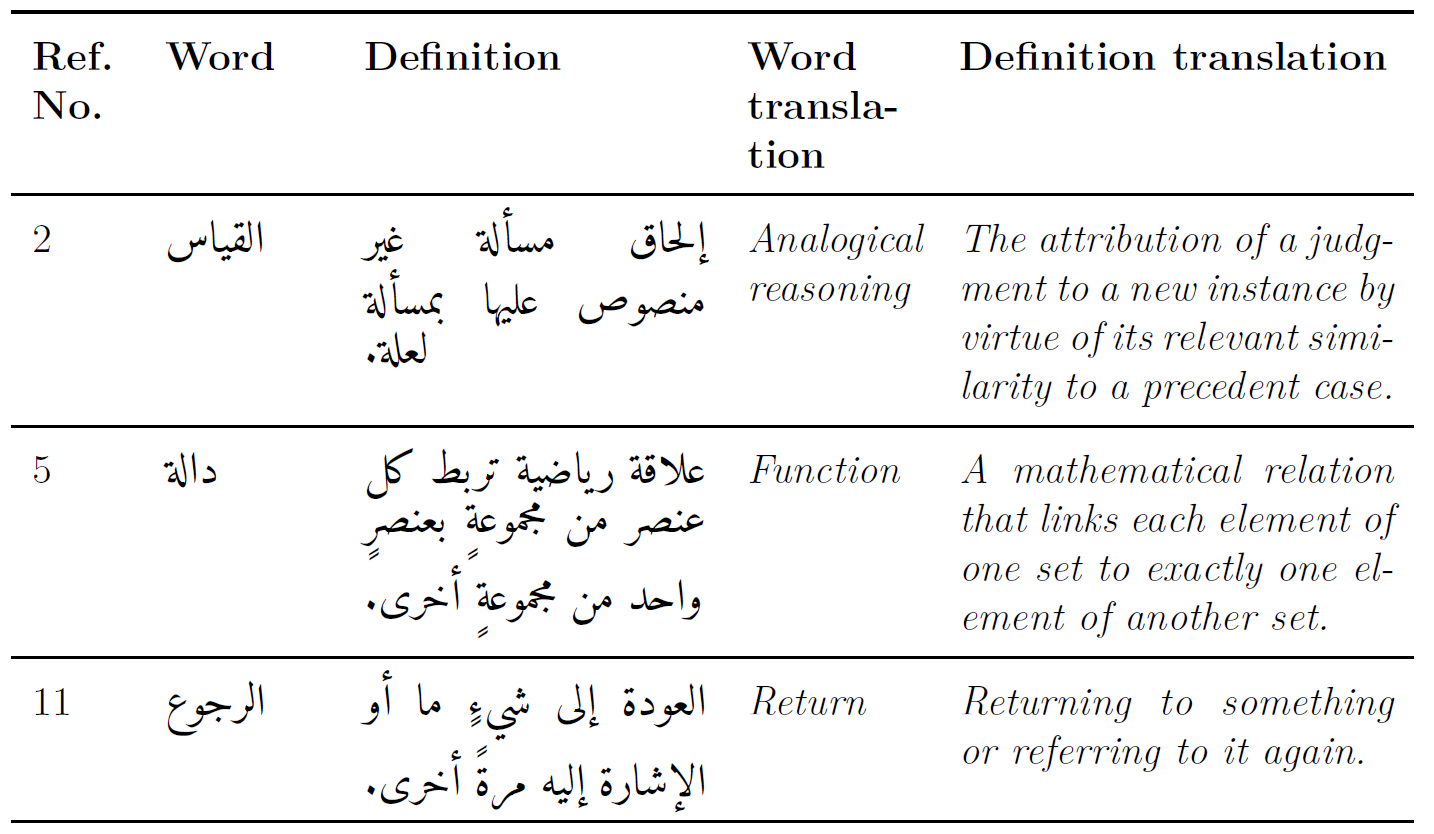}
\caption{Sample Arabic words and definitions from the MURAD dataset. Note: The corresponding English translations are given for readability and are not part of the dataset.}
\label{tab:samples}
\end{table}

\section{Technical Validation}

A series of validation procedures was conducted to ensure the structural integrity, linguistic reliability, and domain representativeness of the MURAD dataset. The validation process focused on three key aspects: completeness, linguistic coherence, and balanced coverage across subject domains. Validation was guided by established standards for quality assurance in Arabic NLP corpora and definition-based modeling \cite{AlMatham2023,Alshammari2024,Mickus2022}.

\subsection{Data Integrity and Completeness}

Data integrity was verified through automated quality control scripts designed to identify and remove incomplete or duplicate records. Each record was checked to confirm the presence of a valid Arabic term paired with its corresponding definition. Automated validation steps ensured that every row contained non-null values and valid UTF-8 encoded text.


\subsection{Linguistic Consistency and Variability}

\begin{figure}[h!]
    \centering
    \includegraphics[width=\textwidth]{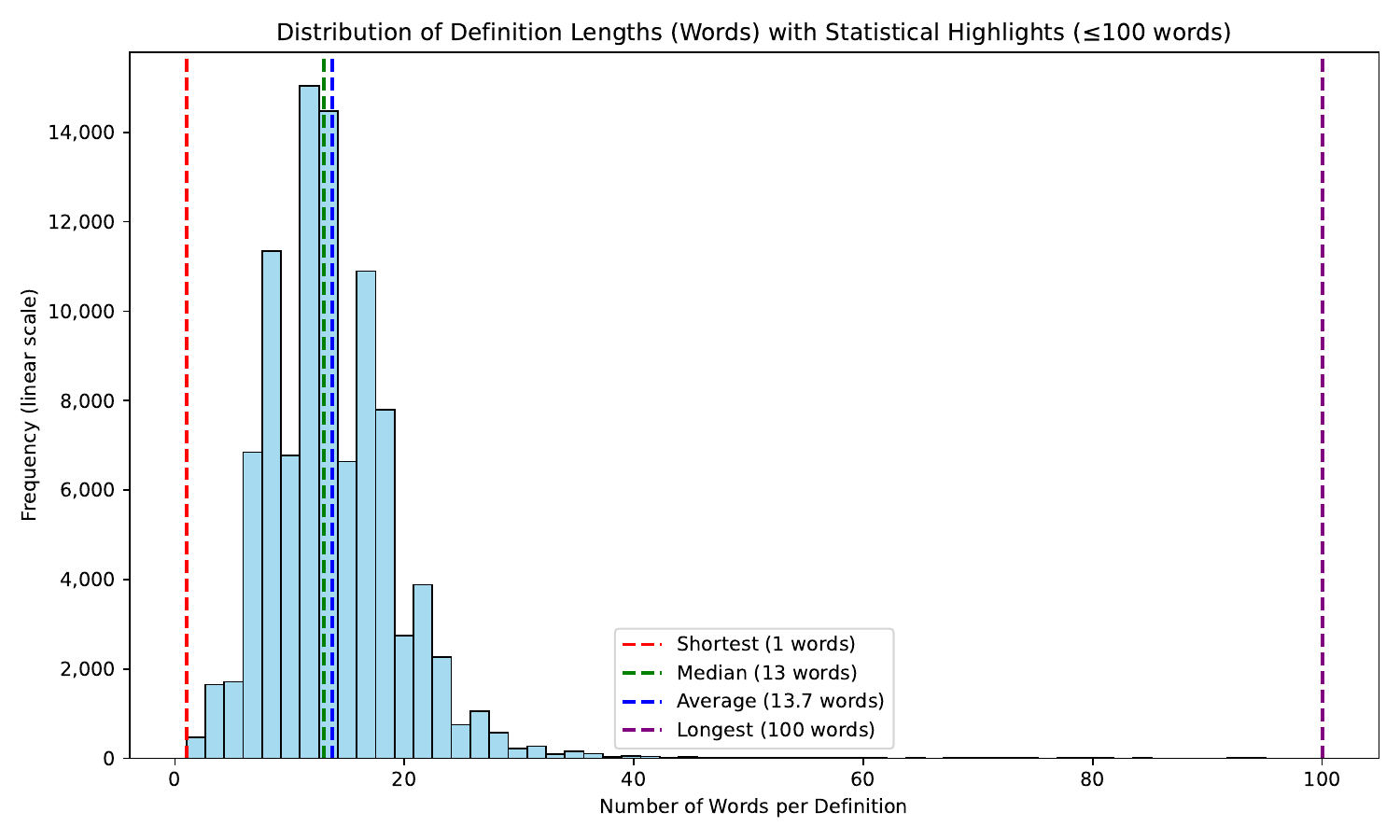}
    \caption{Distribution of definition lengths (in number of words), with statistical highlights indicating the shortest, median, average, and longest entries.}
    \label{fig:lengths}
\end{figure}

Linguistic validation assessed the natural variability and coherence of definition lengths by computing word counts for all entries and deriving descriptive statistics. As shown in Figure~\ref{fig:lengths}, most definitions contain between 10 and 20 words, with a median of 13 and a mean of 13.8 words. The distribution follows a right-skewed pattern, with a small number of extended definitions reaching up to 100 words, reflecting authentic variation in definitional styles across source materials.

This observed distribution aligns with the expected properties of Arabic lexicographic data, in which concise technical definitions coexist with more detailed explanatory entries. Similar variability has been noted in multilingual resources \cite{Mickus2022}, suggesting that definition length is a meaningful linguistic feature rather than a source of noise. The consistency of definition lengths, combined with the absence of extreme outliers, indicates a linguistically stable corpus suitable for definition modeling and semantic retrieval applications.

\subsection{Domain Balance and Representativeness}

\begin{figure}[h!]
    \centering
    \includegraphics[width=\textwidth]{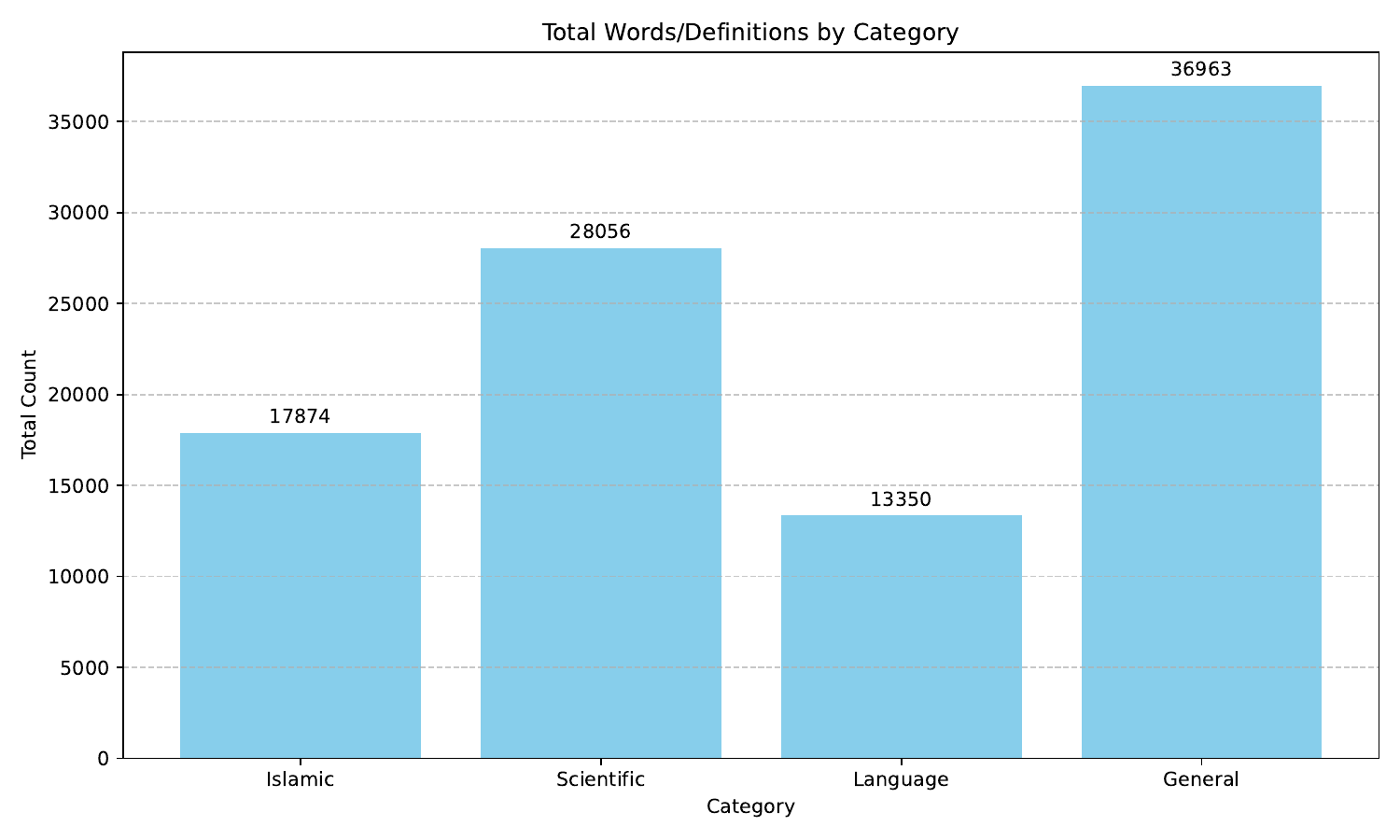}
    \caption{Domain-level distribution of total definitions across the four major source categories.}
    \label{fig:categories}
\end{figure}

Domain-level validation evaluated lexical coverage across four primary thematic categories: \textit{Islamic}, \textit{Scientific}, \textit{Linguistic}, and \textit{General}. Each entry's domain classification was determined based on its source of origin, as detailed in Table~\ref{tab:sources}. Figure~\ref{fig:categories} presents the distribution of definitions across these domains.

The dataset demonstrates a diverse domain distribution, with General terms constituting the largest portion at 38.41\%, followed by Scientific terms at 29.15\%, Islamic terms at 18.57\%, and Linguistic terms at 13.87\% of the total content. This multi-domain composition ensures that the dataset captures a broad spectrum of linguistic and conceptual knowledge, spanning classical Islamic scholarship, modern scientific terminology, linguistic concepts, and general-purpose vocabulary. The predominance of General and Scientific terms reflects the comprehensive nature of modern Arabic lexicography, while the substantial representation of Islamic terminology (18.57\%) preserves the cultural and religious foundations integral to Arabic language resources. This proportional diversity facilitates robust generalization across diverse Arabic NLP tasks, consistent with prior domain-inclusive dataset designs such as KSAA-RD~\cite{AlMatham2023,Alshammari2024}.

\subsection{Reproducibility and Transparency}

All validation and analysis procedures were implemented using open-source tools, including \texttt{pandas}, \texttt{numpy}, and \texttt{matplotlib}. The corresponding scripts are available within the dataset GitHub repository\footnote{\url{https://github.com/riotu-lab/RD-creation-library-RDCL}\label{my-note}} to enable independent verification and reuse. All figures and statistics presented in this section were generated directly from the released dataset without additional filtering or modification, ensuring full reproducibility of the reported results.

This transparent validation framework follows open-data practices established in previous Arabic NLP research \cite{AlMatham2023,Alshammari2024,Sibaee2025}, supporting the FAIR principles (Findable, Accessible, Interoperable, Reusable)~\cite{jacobsen2020fair,barker2022introducing} and facilitating future benchmarking, auditing, and dataset expansion by the research community.

Together, these validation procedures confirm that the MURAD dataset maintains structural completeness, linguistic coherence, and balanced domain coverage. The open and reproducible validation framework ensures long-term usability and trustworthiness, providing a solid foundation for future research in Arabic lexical semantics and definition-based modeling.

\section{Data Availability}

The complete MURAD dataset is publicly available on the Hugging Face Datasets platform\footnote{\url{https://huggingface.co/datasets/riotu-lab/MURAD}}, released for research and educational use.

All data files are distributed under an open research license (CC BY 4.0) and are intended for non-commercial academic and research use. The repository includes the full dataset in CSV format and detailed documentation. The structured release supports the FAIR data principles and is permanently archived for reproducible research.

\section{Code Availability}

All code used for data extraction, processing, validation, and analysis is publicly available through the RIOTU Lab GitHub repository\footref{my-note}. This library provides a complete workflow for extracting text from PDF documents, processing Arabic content, and creating term-definition datasets. It is designed for researchers working with Arabic texts and PDF documents.

\section*{Author Contributions}

Serry Sibaee led the project, conceptualized the study, and was primarily responsible for writing the manuscript. Nadia Sibai and Yara Farouk contributed to manuscript writing and assisted in expanding and curating the dataset. Yasser Al Habashi developed the dataset and provided the technical background and implementation details. Adel Ammar critically reviewed the manuscript, checked the data, and contributed to editing and refinement. 
Sawsan AlHalawani critically reviewed the manuscript and contributed to editing and refinement. Wadii Boulila supervised the research, provided strategic guidance, and oversaw the overall scientific direction of the work. All authors reviewed and approved the final manuscript.

\section*{Competing Interests}
The authors declare that they have no known competing financial interests or personal relationships that could have influenced the work reported in this paper.

\section*{Acknowledgements}
The authors would like to thank Prince Sultan University for their support.

\bibliographystyle{unsrt}
\bibliography{references2}

\end{document}